# Dual-Stream Global-Local Feature Collaborative Representation Network for Scene Classification of Mining Area

1st Shuqi Fan
School of Computer Science
China University of Geosciences
Wuhan, China
fanshuqi@cug.edu.cn

2nd Haoyi Wang
School of Computer Science
China University of Geosciences
Wuhan, China
wanghaoyi.d@gmail.com

3rd Xianju Li *
School of Computer Science
China University of Geosciences
Wuhan, China
ddwhlxj@cug.edu.cn

*Abstract*—The scene classification of mining areas provides accurate foundational data to support geological environment monitoring and resource development planning. This study fuses multi-source data to construct a multi-modal mine land cover scene classification dataset. A significant challenge in mining area classification lies in the complex spatial layout and multi-scale characteristics of these regions. By extracting global and local features, it becomes possible to comprehensively reflect the spatial distribution and overall arrangement of different landforms, thereby enabling a more accurate capture of the holistic characteristics of mining scenes. We propose a dual-branch fusion model utilizing collaborative representation to decompose global features into a set of key semantic vectors. This model comprises three key components: (1) Multi-scale Global Transformer Branch: This branch leverages adjacent large-scale features to generate global channel attention features for small-scale features, effectively capturing the multi-scale feature relationships inherent in mining areas. (2) Local Enhancement Collaborative Representation Branch: This branch refines the attention weights by leveraging local features and reconstructed key semantic sets, ensuring that the local context and detailed characteristics of the mining area are effectively integrated. This enhances the model's sensitivity to fine-grained spatial variations within the mining environment. (3) Dual-Branch Deep Feature Fusion Module: This module fuses the complementary features of the two branches to incorporate more scene information. This fusion strengthens the model's ability to distinguish and classify complex mining landscapes. Finally, this study employs multi-loss computation to ensure a balanced integration of the modules. The overall accuracy of this model is 83.63%, which outperforms other comparative models. Additionally, it achieves the best performance across all other evaluation metrics. The experimental results demonstrate the effectiveness of the proposed dataset and model for classifying mining areas.

## I. Introduction

The mining areas and their surrounding forest zones, agricultural lands, and other surface cover types form a complex geological environment. Mining operations often lead to environmental changes, ecosystem damage, and safety risks, making the monitoring of land cover types in mining areas crucial for sustainable development [1-4]. With the rapid development of remote sensing technology, the application of high-resolution satellite imagery, LiDAR data, and other advanced remote sensing data has gradually become mainstream in mine monitoring and management. Driven by this trend, mining remote sensing scene classification technology has gradually gained attention [5-9].

*Corresponding author

Commonly used high-resolution image classification datasets, such as the UC Merced Land-Use (UCM) dataset [10], Aerial Image Dataset (AID) [11], WHU-RS19 dataset [12], and NWPU-RESISC45 dataset [13], are mainly composed of three spectral bands: red, green, and blue (RGB). However, these optical images are susceptible to interference from factors such as lighting, weather, and cloud cover, significantly limiting the accuracy of scene classification [14]. Additionally, previous studies have pointed out that topographical features are crucial for mining areas [15-17]. Therefore, it is necessary to develop a mine surface cover scene classification dataset that incorporates multiple data modalities to address these challenges more effectively. The irregularity and substantial size differences in the surface features of mining regions exacerbate the difficulty of classification. Moreover, the varying topographical features across different mining areas add additional layers of complexity to the task, making it even more challenging.

The purpose of land cover scene classification in remote sensing images is to accurately label the images using predefined land cover semantic categories [18]. This classification method focuses on accurately assigning pixels or regions in remote sensing images to their respective land cover types. In recent years, deep learning models have become a powerful solution, thanks to the availability of large annotated datasets, advancements in machine learning, and improvements in computational resources [18]. Especially convolutional neural networks (CNNs), due to their remarkable ability to extract spatial and semantic information from high-resolution remote sensing images, have been widely applied and studied in relevant fields [19-21]. Existing studies have confirmed that, due to the scale variation of target objects in remote sensing images, adopting a multi-scale feature extraction strategy is an effective approach to improve classification performance. For instance, Liu et al. [22] proposed a dual-branch network architecture consisting of a fixed-scale network and a variable-scale network to capture feature information at different scales. Wang et al. [23] constructed a global-local dual-stream network based on a structured key area localization strategy. Tian et al. [24] introduced a multi-scale dense network incorporating the SE attention module [25] to enhance classification accuracy.

In the feature extraction process, considering the diversity of objects and features in remote sensing images, selectively focusing on key regions while ignoring irrelevant parts is crucial. Zhao et al. [26] extended the DenseNet-101 network, pre-trained on the ImageNet dataset, by adding a channel-spatial attention module [27] after each residual dense block to obtain more effective feature representations. Ji et al. [28] proposed an attention network based on the VGG-VD16 architecture, which can locate discriminative regions at three different scales. Zhang et al. [29] introduced a multi-scale attention network with a ResNet backbone, which integrates multi-scale modules and channel-position attention modules to extract discriminative regions. Zhao et al. [30] designed an enhanced attention module, employing two convolutional branches with different depth dilation rates to enhance the receptive field while not increasing the number of parameters. Ouyang et al. [31] introduced a multi-modal terrain recognition framework that combines contextual geological features with a channel attention module.

In the field of land cover scene classification, although CNN-based architectures have achieved significant success, these methods cannot comprehensively capture global contextual features. In contrast, Transformer architectures have demonstrated exceptional performance in capturing long-range contextual features [32-34]. For example, Bazi et al. [34] successfully applied the vision transformer (ViT) to remote sensing scene classification. Bashmal et al. [35] employed the data-efficient vision transformer, which achieved impressive results on small datasets through knowledge distillation. Bi et al. [36] combined ViT with supervised contrastive learning (CL) [37] to form the ViT-CL model, further enhancing classification performance.

CNN has an advantage in preserving local information, while ViT is proficient in learning long-range contextual relationships. Therefore, hybrid approaches that combine CNN and Transformer can fully use the advantages of both architectures. For example, Xu et al. [38] have combined ViT with CNN to use the rich contextual information extracted by ViT. Tang et al. [39] proposed an efficient multi-scale transformer and cross-level attention learning method. Wang et al. [40] used the pre-trained Swin Transformer [41] to capture features at multiple layers and connected features through a patch merging strategy, excluding the last block, forming the swin transformer block. Guo et al. [42] integrated channel-space attention into ViT [32], creating the channel-space attention transformer architecture.

This study employs high-resolution imagery from the Gaofen-6, Gaofen-3, and Ziyuan-3 satellites to construct a multimodal land cover scene classification dataset for mining areas. This effort aims to compensate for the current lack of multimodal datasets for mining area land cover. Moreover, a key challenge in mining area classification lies in the complex spatial layout and multi-scale characteristics of these regions. To address these limitations, we propose a dual-stream global-local feature collaborative representation network (DFCRNet). This network combines the advantages of Transformer and CNN in a dual-branch feature extractor. The innovations of this study are as follows:

- In this study, a new mine remote sensing scene dataset was constructed using images from the Gaofen-6 and Gaofen-3 satellites. This dataset is a meticulously designed multimodal collection, integrating multispectral, synthetic aperture radar (SAR), digital elevation model (DEM), and topographical data to assist in classification.

- To address the issue of small target features vanishing in deep networks, we introduce a collaborative dictionary learning module (CDLM) in the multi-scale global Transformer branch to construct a shared dictionary that can collaboratively represent the sample features. In the local enhancement collaborative representation branch, we input multi-layer features into the local feature enhancement module (LFEM) and integrate the dictionary learned by the CDLM. By evaluating the unique contribution of local features to global semantics, we enhance the attention to key regions in the multi-layer feature maps.

- For prominent objects in complex scenes, the extracted semantic features can vary significantly due to differences in size, shape, color, lighting, and imaging conditions. We introduce a deep feature weighted fusion module (DFWFM) to balance global semantics and local information.

## II. DATASET

### A. Study Area Location and Multi-source Remote Sensing Data

This study focuses on the V3 region of Hubei Province, located within the Yangtze River Three Gorges geological environment subregion in China. The geographic coordinates of the study area range from 110°6' E to 112°8' E, and 30°15' N to 31°37' N, as illustrated in Fig. 1. Constructing a dataset covering a larger geographic area would require significantly higher costs. By selecting the V3 region as a pilot study area, this research establishes a foundation for future expansion.

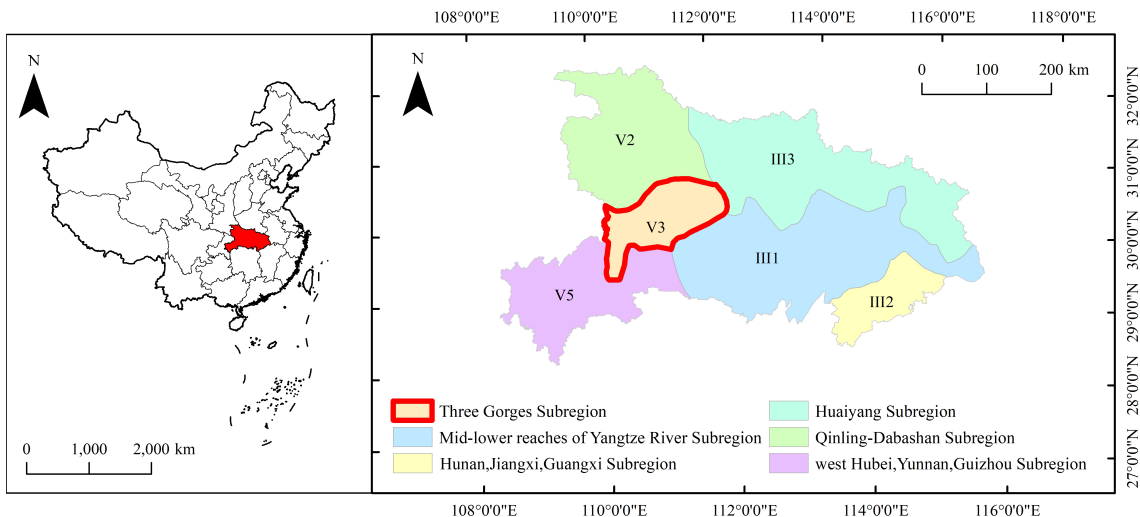

Fig. 1. Location of the study area.

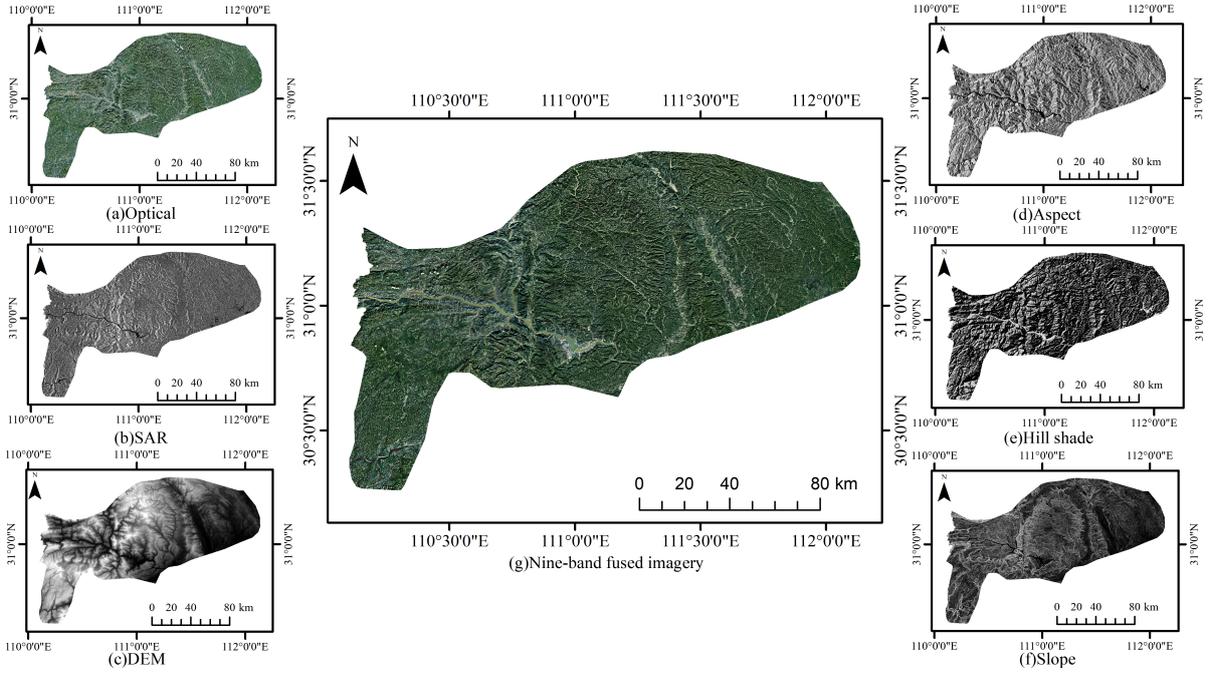

Fig. 2. The multi-source data and nine-band imagery of the study are.

In this study, we integrated multiple remote sensing data sources to ensure the comprehensiveness and accuracy of the information for the study area. Specifically, we used GF-6 satellite imagery to obtain RGB and near-infrared data. First, orthorectification was performed on both panchromatic and multispectral datasets from the GF-6 satellite, followed by geometric registration of the multispectral data. The datasets were then fused to generate a four-band optical image with 2-meter spatial resolution. Given the limitations of optical sensors in terms of penetration ability, as well as their susceptibility to cloud cover and vegetation, we incorporated SAR data from the GF-3 satellite, which has a spatial resolution of 5 meters and enhances penetration and sensitivity to surface roughness. Additionally, considering the impact of open-pit mining on terrain features, we obtained DEM data from ZiYuan-3 stereo imagery (10-meter resolution) to extract key topographic features such as slope, aspect, and hill shade.

Due to the differences in spatial resolution, we resampled the SAR data, DEM data, and their derived topographic feature data to a 2-meter resolution to match the optical imagery. After geometric correction and resampling, we performed multi-source information fusion, integrating optical imagery, SAR data, DEM data, and topographic features to construct a 9-band remote sensing image dataset. Fig. 2 illustrates the multi-source data and the fused imagery of the study area.

### B. Construction of the Dataset

In this study, we manually interpreted the mining targets in the study area through visual analysis and combined this with the national-level 1-meter resolution land cover map of China [43]. The images were cropped into 256×256 pixel patches using a non-overlapping sliding window technique, following a left-to-right, top-to-bottom sequence.

Given the widespread forest cover in the study area and the relative scarcity of other categories, we randomly selected a proportion of images from the tree cover scenes to avoid data redundancy. Table I presents the description of the classification system and the number of images in the dataset.

TABLE I. DESCRIPTION OF THE DATASET CLASSIFICATION SYSTEM AND NUMBER OF IMAGES IN THE DATASET

| Lable | Name | Definition | Number |
|---|---|---|---|
| 0 | Mine | It refers to the area where mining operations are carried out. | 672 |
| 1 | Tree cover | Areas covered by trees generally have larger crowns and are higher than 5 m. | 800 |
| 2 | Cropland | Areas covered by arable land and human-planted crops not at tree height. | 204 |
| 3 | Water | Areas covered by water for a long period, including oceans, naturally formed water bodies, and artificially formed water bodies. | 869 |

The multi-modal mining land cover scene classification dataset created in this study consists of 2,545 images, each containing 9 bands. The dataset covers four types of mining land cover: mine, tree cover, cropland, and water. Subsequently, the dataset was divided into training, validation, and test sets in a 6:2:2 ratio. Fig. 3 presents example images from each category in the dataset.

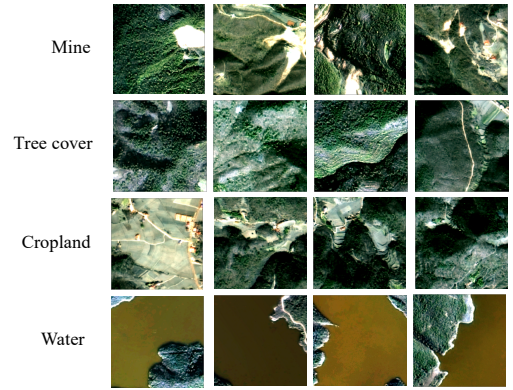

Fig. 3. Example images for each category in the dataset.

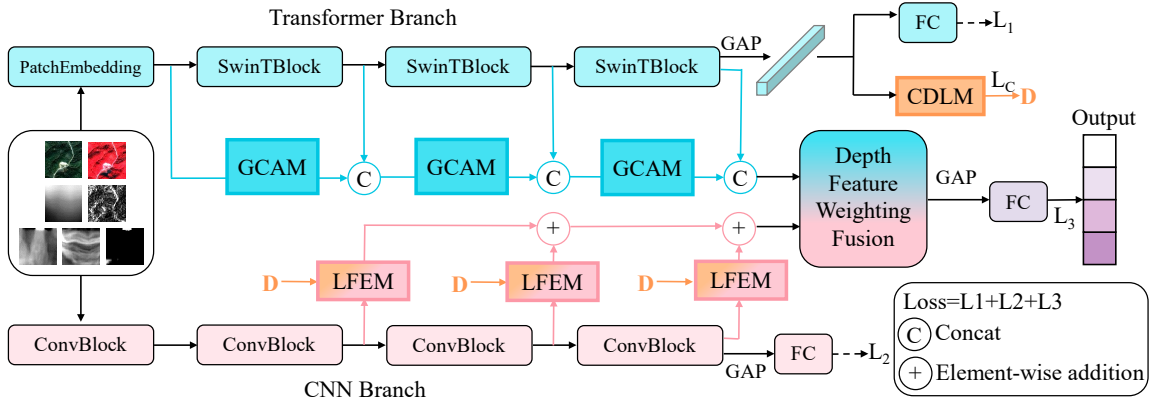

Fig. 4. DFCRNet network architecture.

## III. METHODS

We designed DFCRNet, which combines the advantages of CNNs and Transformers, as shown in Fig. 4. The framework incorporates four modules to enhance classification accuracy. These modules are introduced to improve feature representation and fusion across the Transformer and CNN branches. The detailed description of each module will be provided in the following sections.

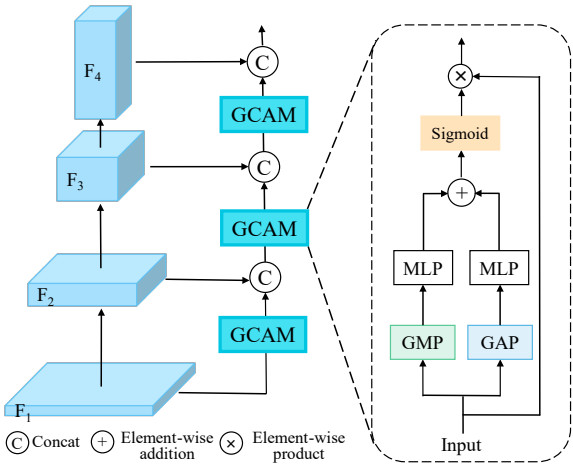

Fig. 5. Multi-scale global transformer branch.

### A. Multi-Scale Global Transformer Branch

The Transformer branch utilizes a global attention mechanism to connect all pixel information, but this may introduce background noise. The introduction of multi-scale features helps expand the information flow. We have introduced GCAM at four stages of the Transformer branch, leveraging channel attention to enhance the representational capacity of important feature channels. As shown in Fig. 5, four feature maps $F_1$, $F_2$, $F_3$, and $F_4$ containing multi-scale semantic information are generated by the Swin Transformer Block. Then, a bottom-up structure is applied to fuse these feature maps, with downsampling applied to the larger-scale feature maps during the fusion process. For a feature map $F$ of size $H \times W \times C$, global average pooling and global max pooling are first applied to $F$, resulting in two channel maps of size $1 \times 1 \times C$. Subsequently, a multi-layer perceptron (MLP) with hidden layers is applied to generate the weight channels for the features. To reduce the parameter overhead, the size of the hidden layers is set to $C/r$, where $r$ is the reduction factor. To effectively combine these two types of features, their respective results are fused through element-wise addition. Finally, the feature map is improved by multiplying it with the channel weights, where the weight coefficients are obtained through a sigmoid activation function. The formula for GCAM is as follows:

$$M_C(F) = \sigma\left(\text{MLP}(\text{GMP}(F)) + \text{MLP}(\text{GAP}(F))\right) \times F, \quad (1)$$

where $\sigma$ is the sigmoid activation function, GMP and GAP represent global max pooling and global average pooling, respectively, and $F$ is the input feature map.

### B. Local Enhancement Collaborative Representation Branch

As shown in Fig. 4, the Transformer branch captures contextual information and reconstructs global semantic features by optimizing a low-redundancy dictionary through the CDLM. CDLM enhances feature sparsity and improves class separability, which helps distinguish mining areas with similar spectral characteristics but different spatial patterns. In the CNN branch, to prevent the loss of small-object features, we transform multi-layer features into a unified spatial and channel dimension consistent with the final layer. These refined features, along with the dictionary learned from CDLM, are utilized in LFEM to enhance fine-grained spatial details, enabling better differentiation of small-scale mining sites from other land categories. Finally, multi-layer attention feature maps are fused and used as the output of the CNN branch.

*1) Collaborative Dictionary Learning Module:* CDLM aims to learn an optimal, sparse representation dictionary $D=[d_1, d_2, \cdots, d_K]$ from a large set of data samples, where $K$ denotes the number of atoms in the dictionary, which corresponds to the number of categories in this study. As shown in Fig. 6, given a randomly initialized dictionary $D$ and a feature vector $x$, we obtain a coefficient vector $\hat{s}$ that represents the projection of the data onto the dictionary basis functions. This can be defined as follows:

$$\hat{s} = ((WD)^T(WD) + \lambda I)^{-1}(WD)^T x, \quad (2)$$

where $W$ is a learnable linear transformation matrix that maps the coefficient space to the data space; $I$ is the identity matrix, ensuring the consistency of the transformation; and $\lambda$ is a hyperparameter controlling regularization strength. Our goal is to approximate the data sample $x$ through a linear combination of the atoms (i.e., column vectors) in D. We optimize the dictionary $D$ by minimizing the distance loss $L_C$ between the original data $x$ and the reconstructed feature vector $y$ from the coefficient vector $\hat{s}$. This distance is typically measured using the L2-norm (i.e., Euclidean distance), and the loss function $L_C$ is expressed as follows:

$$L_C = \left\| \frac{x}{\|x\|_2} - \frac{y}{\|y\|_2} \right\|_2^2. \quad (3)$$

$L_C$ normalizes both $x$ and $y$ to ensure numerical stability during the optimization process. To stabilize the learning process of the dictionary $D$, the loss function $L_C$ avoids backpropagation through $x$. The dictionary $D$ is typically initialized randomly to allow the algorithm to explore the solution space from different perspectives.

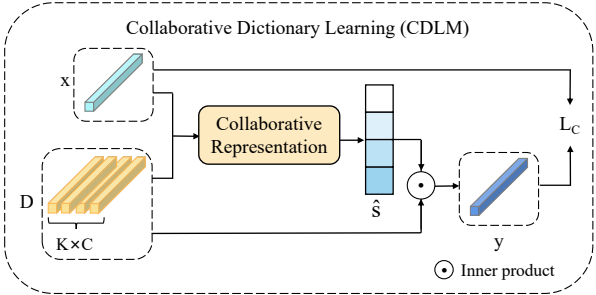

Fig. 6. Collaborative dictionary learning module.

*2) Local Feature Enhancement Module:* LFEM enhances the key local regions within the feature map by integrating the $D$ learned by the CDLM. As shown in Fig. 7, we reshape the input feature map $F$ from $H \times W \times C$ to $N \times C$, where $N$ is the total number of elements after flattening the feature map. Concurrently, we combine $W$, $\hat{s}$, and $D$ from CDLM to compute the reconstruction of the key semantic set $Z=[z_1,z_2,\cdots,z_K]$. Next, fully connected (FC) layers are applied to both the reshaped feature map $F'$ and the reconstructed key semantic set $Z$. To assess the correlation between $F'$ and $Z$, we calculate their inner product, yielding the result $T=[t_1,t_2,\cdots,t_N]$, $T \in R^{N \times K}$. We then compute the column mean of $T$ to measure the contribution of each feature vector to the global semantics. These contributions are normalized using the Softmax function to obtain the attention coefficient vector, ensuring the sum of all attention coefficients equals 1. Finally, we perform an element-wise product between the normalized attention coefficient vector and the original feature map, while introducing a residual connection to enhance the model's learning capacity and generalization ability.

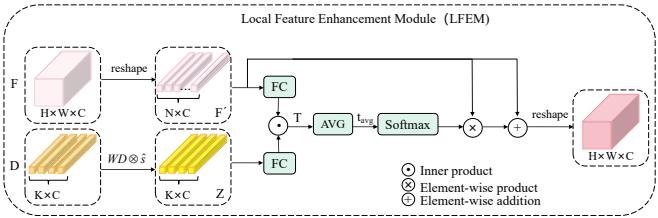

Fig. 7. Local feature enhancement module.

*C. Dual-Branch Deep Feature Weighted Fusion Module*

We fuse features from two branches to integrate more scene information and enhance classification accuracy. We propose DFWFM to balance the global and local features of salient regions. DFWFM integrates both global context and local detail information, using a feature weighting strategy to dynamically adjust the importance of the features. The architecture of this module is shown in Fig. 8. The blue feature maps represent the features from the Transformer branch, while the red feature maps represent the features from the CNN branch. First, the spatial information of the feature maps $F_1$ and $F_2$ from the two branches is aggregated using global average pooling and global max pooling, respectively. After applying the Sigmoid activation function to obtain the weights, these are multiplied with the original input feature maps $F_1$ and $F_2$ to obtain more discriminative features. The two branches are added together to form a third branch, with each of the three branches passing through 3×3 depthwise separable convolutions. The third branch is then concatenated with the original two branches, and the concatenated feature map is passed through a 1×1 convolution for further learning. This result is then added to the feature map before convolution via a residual connection. Finally, the results from the two branches are concatenated to form the output of the fusion module.

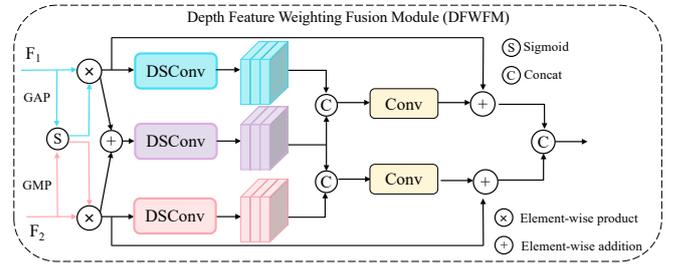

Fig. 8. Deep feature weighted fusion module.

IV. RESULTS

*A. Experimental Settings*

The experiments are conducted in an environment running the Windows 10 operating system, and the scene classification algorithm is implemented using the PyTorch framework. The hardware configuration includes 64GB of memory, an AMD EPYC processor, and an NVIDIA RTX A5000 GPU with 24GB of video memory.

Before the experiments, all images undergo standardized preprocessing. The learning rate is set to 0.0001, and the Adam optimizer is used for network training. For different algorithm implementations, the batch sizes are set to 32 and 16, respectively, and each experiment involves 100 training iterations. Each experiment is repeated 5 times, with the final results being the average and standard deviation to ensure the reliability of the data.

To provide a benchmark for future research, we compared the proposed model with several classical models (including VGGNet-16[44], ResNet-101[45], DenseNet-121[46], and ConvNeXt[47]), Transformer-based models (Swin Transformer[41]), and three other recently proposed classification models on a multi-modal land cover scene classification dataset for mining areas.

We used the overall accuracy (OA), precision, recall, F1-Macro, and Kappa coefficient as key metrics to evaluate the model's performance. Given the imbalanced class distribution in the dataset, all evaluation metrics are reported using macro-averaging to ensure a fair assessment across all categories.

*B. Experimental Results*

Table II provides a quantitative comparison of our method with other approaches, with the highest accuracy highlighted in bold. The results show that DFCRNet outperforms all other models across all evaluation metrics. The results indicate that DFCRNet effectively combines the strengths of both Transformer and CNN architectures, capturing local fine-grained representations while modeling global relationships.

Table III presents the detailed experimental results of different models on each category of the test set, with the highest accuracy in bold and the second-highest underlined. Our model achieves favorable performance across all categories. Notably, in the key category of mines, our model achieves the highest accuracy. This performance improvement can be attributed to the effective combination of the dual-branch structure, which allows the model to leverage the different feature representation capabilities of each branch.

TABLE II. EXPERIMENTAL RESULTS OF DIFFERENT MODELS

| Methods | OA(%) | Precision(%) | Recall(%) | F1-Macro(%) | Kappa(%) |
|---|---|---|---|---|---|
| VGGNet-16[44] | 79.38±0.34 | 79.35±0.31 | 77.97±0.15 | 77.55±0.26 | 70.7±0.14 |
| ResNet-101[45] | 80.6±0.45 | 79.36±0.22 | 76.42±0.15 | 77.36±0.34 | 70.34±0.42 |
| DenseNet-121[46] | 82.49±0.18 | 79.37±0.19 | 80.48±0.17 | 79.22±0.41 | 75.36±0.23 |
| ConvNeXt-S[47] | 79.4±0.17 | 79.34±0.45 | 75.14±0.45 | 76.47±0.27 | 70.61±0.11 |
| SwinTransformer-T[41] | 81.26±0.15 | 80.71±0.12 | 78.69±0.15 | 79.42±0.36 | 73.43±0.2 |
| SAGN[48] | 81.36±0.25 | 80.58±0.1 | 79.39±0.07 | 79.44±0.37 | 73.52±0.37 |
| DBGA[49] | 78.5±0.36 | 76.2±0.33 | 78.62±0.38 | 75.73±0.39 | 69.66±0.4 |
| CDLNet[50] | 82.15±0.18 | 80.45±0.1 | 80.42±0.15 | 79.45±0.37 | 75.45±0.23 |
| Ours | **83.63±0.16** | **82.31±0.12** | **81.31±0.12** | **81.71±0.24** | **76.17±0.16** |

TABLE III. EXPERIMENTAL RESULTS OF EACH CATEGORY ACCURACY FOR DIFFERENT MODELS

| Methods | Mine (%) | Tree Cover(%) | Cropland (%) | Water (%) | OA (%) |
|---|---|---|---|---|---|
| VGGNet-16[44] | 68.83 | 74.14 | 81.82 | 94.54 | 79.38 |
| ResNet-101[45] | 67.72 | 73.46 | 65.19 | 97.13 | 80.6 |
| DenseNet-121[46] | 69.44 | 77.42 | 75.68 | 96.59 | 82.49 |
| ConvNeXt-S[44] | 64.96 | 72.33 | 84.38 | 96.09 | 79.4 |
| SwinTransformer-T[41] | 67.61 | 74.36 | 79.49 | 96.65 | 81.26 |
| SAGN[48] | 66.67 | 73.89 | 82.76 | 94.54 | 81.36 |
| DBGA[49] | 63.64 | 69.46 | 77.14 | 95.95 | 78.5 |
| CDLNet[50] | 69.83 | 77.33 | 79.68 | 95.09 | 81.49 |
| Ours | **71.24** | **80.28** | **85.71** | **97.18** | **83.63** |

Compared to the accuracy of the "Water" category, the accuracy of the other three categories is relatively lower. Analysis reveals that the "Mine" and "Tree Cover" categories in the dataset are prone to misclassification, primarily because mining features are predominantly distributed in mountainous areas covered by forests, and the sample sizes of these two categories differ significantly. Similarly, the "Cropland" category is often misclassified as "Mine" because farmland is typically located in mountain valleys near human settlements, and the limited number of images for this category affects the model's recognition capability.

## V. DISCUSSION

### A. Ablation Experiment

We designed ablation experiments to analyze the contribution of different modules in DFCRNet. According to the data in Table IV, model 6 achieves the highest values in OA, precision, recall, F-1 Macro, and Kappa. Model 2 shows increased OA and precision compared to model 1, indicating that the multi-scale GCAM effectively captures important channel information. By comparing the results of model 1 and model 3, it is clear that the incorporation of CDLM and LFEM improves the classification accuracy of categories. Finally, when combined with other modules, DFWFM helps integrate diverse feature representations.

TABLE IV. ABLATION EXPERIMENT RESULTS FOR EACH MODULE

| Model | GCAM | CDLM+LFEM | DFF | OA(%) | Precision(%) | Recall(%) | F1-Macro(%) | Kappa(%) |
|---|---|---|---|---|---|---|---|---|
| 1 | | | | 80.01±0.49 | 80.41±0.35 | 78.35±0.31 | 78.53±0.37 | 72.21±0.25 |
| 2 | √ | | | 81.46±0.28 | 80.63±0.25 | 78.71±0.16 | 79.57±0.23 | 73.7±0.16 |
| 3 | | √ | | 81.54±0.41 | 80.17±0.27 | 79.17±0.15 | 80.12±0.2 | 73.76±0.42 |
| 4 | | | √ | 81.26±0.19 | 81.57±0.25 | 79.27±0.23 | 80.95±0.35 | 73.48±0.31 |
| 5 | √ | √ | | 82.25±0.2 | 82.03±0.33 | 79.87±0.19 | 80.78±0.32 | 74.85±0.21 |
| 6 | √ | √ | √ | **83.63±0.16** | **82.31±0.12** | **81.31±0.12** | **81.71±0.24** | **76.17±0.16** |

### B. Different Attention Modules

To evaluate the specific impact of different attention modules on the experimental results, this study integrates various attention mechanisms within the same deep learning architecture. We selected several representative attention modules and applied them to the same network framework. As shown in Table V, our CDLM+LFEM attention module achieved the best accuracy. This is because the combination of the CDLM and LFEM modules simultaneously leverages both global and local features. CDLM reconstructs long-range contextual global semantic information in the image to obtain more precise attention weights, which then guide LFEM in enhancing key local regions within the feature map.

TABLE V. COMPARISON OF DIFFERENT ATTENTION MODULES BASED ON THE SAME ARCHITECTURE

| Attention Module | OA(%) | Precision (%) | F1-Macro(%) | Kappa(%) |
|---|---|---|---|---|
| SE-Net[25] | 82.26±0.19 | 82.57±0.07 | 79.53±0.3 | 73.91±0.32 |
| ECA[51] | 80.25±0.2 | 79.03±0.33 | 78.78±0.32 | 74.85±0.21 |
| CBAM[27] | 83.04±0.22 | 81.33±0.12 | 81.19±0.14 | 75.88±0.26 |
| CDLM+LFEM | **83.63**±0.16 | **82.31**±0.12 | **81.71**±0.24 | **76.17**±0.16 |

*C. Limitations*

This study has several limitations. First, the dataset from Hubei's V3 region has fewer mining and cropland samples than dominant classes like forests, causing class imbalance that affects classification, especially for mining areas. Also, manual annotation based on visual interpretation is subjective and error-prone.

Second, real-world mining environments are highly dynamic, influenced by seasonal changes, ongoing mining activities, and various noise factors, including abnormal samples. However, the current study does not fully explore the model's robustness and adaptability under such complex and evolving conditions, limiting its practical applicability.

Future work will address these challenges by expanding the dataset, thereby increasing the diversity of mining area samples and improving the model's ability to generalize across different geographic contexts. Furthermore, self-supervised learning will be explored to leverage unlabeled data for pretraining, reducing dependence on label quality and quantity while enhancing the model's ability to capture minority class features. Additionally, incorporating temporal information from multi-temporal remote sensing data will enable the model to better capture dynamic changes in mining activities, improving its adaptability to real-world variations.

## VI. CONCLUSION

Land cover classification of mining areas is crucial for environmental assessment. Remote sensing images from different sources can provide more diverse and complementary information. Therefore, this study has developed a multi-modal land cover scene classification dataset for mining areas. To address the challenges posed by irregularly shaped, small-sized, and sparsely distributed objects in mining areas, and to overcome the limitations of single-branch models in fully capturing significant targets in complex environments, we have designed DFCRNet. The key components of DFCRNet are: (1) Multi-scale Global Transformer Branch, which captures long-range contextual information and fuses multi-level features; (2) Local Enhancement Collaborative Representation Branch, which integrates a global contextual dictionary and a local feature enhancement module to prevent the loss of small target information. This branch combines Transformer-based long-range dependencies with CNN-based strong local biases to model global relationships and capture local details; (3) Dual-Branch Deep Feature Fusion Module, which fuses the complementary features of the two branches.

The DFCRNet model proposed in this study demonstrates excellent performance on the multi-modal mining area land cover scene classification dataset, achieving an overall accuracy of 83.63%, which outperforms comparative models and excels across all evaluation metrics. In the future, we plan to further improve the model's efficiency and cross-domain capability, making it more practical and adaptable. Additionally, we will continue to expand and refine our dataset by incorporating data from five additional regions in Hubei, aiming for a larger, more balanced dataset.

ACKNOWLEDGMENT

This study was jointly supported by the Natural Science Foundation of China under Grants 42071430 and U21A2013, the Opening Fund of Key Laboratory of Geological Survey and Evaluation of the Ministry of Education under Grants GLAB2022ZR02 and GLAB2020ZR14. Computation of this study was performed by the High-performance GPU Server (TX321203) Computing Centre of the National Education Field Equipment Renewal and Renovation Loan Financial Subsidy Project of China University of Geosciences, Wuhan.